%% file: Manuscript.tex
\documentclass[lettersize,preprint]{elsarticle}
\usepackage{amsmath,amsfonts}
\usepackage{subfig}
\usepackage{url}
\usepackage{graphicx}
\usepackage{siunitx}

\usepackage[normalem]{ulem}

\usepackage{xcolor}
\usepackage{algpseudocode}
\usepackage{algorithm}

\usepackage{multirow}
\usepackage{fixmath}

\usepackage{tikz}
\usepackage{makecell}
\usepackage{booktabs}
\usepackage{csquotes}
\usepackage{wrapfig}
\usepackage{parskip}
\usepackage[acronym]{glossaries}
\makeglossaries

\input{tikz_settings}

\usepackage{soul}

\makeatletter
\let\AC@footnote\@gobble
\makeatother

\newcommand{\B}{$\mathcal{B}$}
\newcommand{\BS}{$\mathcal{B}^*$}
\newcommand{\D}{$\mathcal{D}$}
\newcommand{\ps}{\emph{p-state}}
\newcommand{\ts}{\emph{t-series}}

\newacronym{AMT}{AMT}{Active blinking Marker Tracking}
\newacronym{IR}{IR}{Infrared}
\newacronym{GNSS}{GNSS}{Global Navigation Satellite System}
\newacronym{MOCAP}{mo-cap}{Motion capture}
\newacronym{MRS}{MRS}{Multi-robot Systems Group}
\newacronym{MAV}{MAV}{Micro-scale Unmanned Aerial Vehicle}
\newacronym{UWB}{UWB}{Ultra-wideband}
\newacronym{GPU}{GPU}{Graphics Processing Unit}
\newacronym{VIO}{VIO}{Visual Inertial Odometry}
\newacronym{UAV}{UAV}{Unmanned Aerial Vehicle}
\newacronym{UGV}{UGV}{Unmanned Ground Vehicle}
\newacronym{UVDAR}{UVDAR}{UltraViolet Direction And Ranging}
\newacronym{RTK}{RTK}{Real-Time Kinematic}
\newacronym{SIFT}{SIFT}{Scale-invariant Feature Transform}
\newacronym{EKF}{EKF}{Extended Kalman Filter}
\newacronym{KF}{KF}{Kalman Filter}
\newacronym{JPDAF}{JPDAF}{Joint Probabilistic Data-association Filter}
\newacronym{wrt}{w.r.t.}{With Respect To}
\newacronym{4DHT}{4DHT}{4D Hough Transform}
\newacronym{UV}{UV}{Ultraviolet}
\newacronym{TX}{TX}{Transmitter}
\newacronym{RX}{RX}{Receiver}
\newacronym{FOV}{FOV}{Field of View}
\newacronym{LED}{LED}{Light-Emitting Diode}
\newacronym{RF}{RF}{Radio Frequency}
\newacronym{OCC}{OCC}{Optical Camera Communication}
\newacronym{DVS}{DVS}{Dynamic Vision System}
\newacronym{CMOS}{CMOS}{Complementary Metal-Oxide-Semiconductor}
\newacronym{LoS}{LoS}{Line of Sight}
\newacronym{OOK}{OOK}{On-Off Keying}
\newacronym{CNN}{CNN}{Convolutional Neural Network}
\newacronym{GM-PHD}{GM-PHD}{Gaussian Mixture Probability Hypothesis Density}
\newacronym{RSS}{RSS}{Resident Set Size}
\newacronym{FAST}{FAST}{Features from Accelerated Segment Test}
\newacronym{CI}{CI}{Confidence interval} 

\journal{Robotics and Autonomous Systems}

\usepackage{hyperref}

\newcommand{\PUBLISHEDIN}{Robotics and Autonomous Systems}
\newcommand{\DOI}{10.1016/j.robot.2025.105175} 
\usepackage[placement=top,vshift=-7]{background}
\SetBgScale{1.0}
\SetBgContents{\PUBLISHEDIN. AUTHOR'S VERSION - DO NOT DISTRIBUTE. \href{https://doi.org/\DOI}{DOI \DOI}}
\SetBgColor{black}
\SetBgAngle{0}
\SetBgOpacity{1.0}

\begin{document}

\begin{frontmatter}

\title{Towards Agile Multi-Robot Systems in the Real World: Fast Onboard Tracking of Active Blinking Markers for Relative Localization\tnoteref{t1}}
\tnotetext[t1]{This work was funded by CTU grant no SGS23/177/OHK3/3T/13, by the Czech Science Foundation (GAČR) under research project no. 23-07517S and by the European Union under the project Robotics and advanced industrial production (reg. no. CZ.02.01.01/00/22\_008/0004590).}
\tnotetext[t2]{Received 28 January 2025, Revised 13 August 2025, Accepted 16 August 2025, Available online 6 September 2025, Version of Record 13 September 2025.}

\author[1]{Tim Lakemann\corref{cor1}}
\ead{lakemtim@fel.cvut.cz}
\author[2,1]{Daniel Bonilla Licea}
\author[1]{Viktor Walter}
\author[1]{Tomáš Báča}
\author[1]{Martin Saska}

\cortext[cor1]{Corresponding author, +420-22435-7255}
\affiliation[1]{organization={Czech Technical University},
            addressline={Karlovo namesti 13}, 
            city={Prague},
            postcode={121 35}, 
            country={Czech Republic}}
\affiliation[2]{organization={Mohammed VI Polytechnic University}, 
country={Morocco}}

\begin{abstract}
A novel onboard tracking approach enabling vision-based relative localization and communication using \gls{AMT} is introduced in this article.
Active blinking markers on multi-robot team members improve the robustness of relative localization for aerial vehicles in tightly coupled multi-robot systems during real-world deployments, while also serving as a resilient communication system.
Traditional tracking algorithms struggle with fast-moving blinking markers due to their intermittent appearance in camera frames and the complexity of associating multiple of these markers across consecutive frames.
\gls{AMT} addresses this by using weighted polynomial regression to predict the future appearance of active blinking markers while accounting for uncertainty in the prediction.
In outdoor experiments, the \gls{AMT} approach outperformed state-of-the-art methods in tracking density, accuracy, and complexity. 
The experimental validation of this novel tracking approach for relative localization and optical communication involved testing motion patterns motivated by our research on agile multi-robot deployment.
\end{abstract}


\begin{keyword}
Visual Tracking, Localization, Multi-Robot Systems, Computer Vision for Automation
\end{keyword}

\end{frontmatter}

\section{Introduction}
Collaborative multi-\gls{UAV} systems offer enhanced redundancy and efficiency compared to single-robot systems, but require accurate mutual localization for collision avoidance and coordinated task execution~\cite{chungSurveyAerialSwarm2018, chenSurveyRobotSwarms2022}.
\gls{GNSS} suffers from limited accuracy and is susceptible to jamming and spoofing~\cite{xuDecentralizedVisualInertialUWBFusion2020a, gaoVIDORobustConsistent2023}, whereas \gls{RTK}-\gls{GNSS} and motion capture systems offer centimeter-level precision at the cost of external infrastructure dependence~\cite{chungSurveyAerialSwarm2018, chenSurveyRobotSwarms2022, wuRealizationRemoteMonitoring2022}.
Onboard \gls{UWB} provides comparable distance accuracy without external infrastructure.
However, it remains susceptible to radio interference and requires anchor networks for full pose estimation~\cite{xuDecentralizedVisualInertialUWBFusion2020a, jiangIndoorOutdoorSeamless2021, queraltaUWBbasedSystemUAV2020, zhouSwarmMicroFlying2022}.

Camera-based mutual relative localization offers a robust and cost-effective solution for multi-robot systems without external infrastructure~\cite{schillingVisionBasedDroneFlocking2021}.
Recent relative localization algorithms often rely on \glspl{CNN}, which impose high computational demands, suffer from illumination sensitivity, and lack cross-platform applicability~\cite{xuDecentralizedVisualInertialUWBFusion2020a, schillingVisionBasedDroneFlocking2021, ohMarkerBasedLocalizationSystem2023}.
Passive markers attached to robot platforms offer a low-cost alternative but require stable lighting and planar surfaces, restricting use on small, agile UAVs~\cite{romero-ramirezSpeededDetectionSquared2018a, krajnikExternalLocalizationSystem2013}.
To overcome these limitations, active markers in the form of \glspl{LED} can be attached to team members, enabling efficient background separation and robust operation in challenging indoor and outdoor environments, while supporting simultaneous communication and relative localization~\cite{stuckeyRealTimeOpticalLocalization2024, whiteOpticalSpatialLocalization2019, teixeiraVIRPEVisualInertialRelative2018, walterUVDARSystemVisual2019}.
By modulating blinking patterns using modulation techniques such as \gls{OOK}, these markers enable both unique identification and \gls{OCC} among team members, supporting decentralized coordination without relying on external infrastructure~\cite{whiteOpticalSpatialLocalization2019, liceaOpticalCommunicationbasedIdentification2023, stuckeyOpticalSpatialLocalization2021, diasOnboardVisionbased3D2016}.
When using agile flying \glspl{UAV} with blinking light sources attached, classical tracking algorithms such as the \gls{KF} with non-uniform sampling time~\cite{plarreKFwithIntermittentObservations2009} can, in principle, handle intermittent appearances caused by occlusions or asynchronous blinking.
However, these approaches depend on consistent data association across frames, which becomes unreliable when tracking multiple indistinguishable blinking light sources in close proximity within the image -- a common scenario in multi-\gls{UAV} systems.
Advanced multi-target tracking methods, such as \gls{JPDAF} combined with the \gls{EKF}, can resolve ambiguous measurement associations but often merge tracks of nearby targets when detections alternate between frames, making them unsuitable for tracking multiple identical, independently blinking light sources in close proximity~\cite{wangMultiTargetVideoTracking2016}.

Recent studies on agile multi-\gls{UAV} teams reach inter-agent speeds up to \SI{2}{\meter\per\second} in object-dense outdoor environments~\cite{zhouSwarmMicroFlying2022} and up to \SI{7.4}{\meter\per\second} indoors with external localization infrastructure~\cite{ryouCooperativeMultiAgentTrajectory2022}.
Even at lower inter-agent speeds, robust tracking algorithms are essential for maintaining reliable \gls{OCC} links and accurate relative localization.

\begin{figure}[t]
  \centering
  \scalebox{1.5}{\input{desert.tex}}
  \caption{Six agile swarm members flying in challenging desert conditions, relying on \glsentryshort{UVDAR}-based relative localization as used in this work.}
  \vspace{-5pt}
  \label{fig:intro:overview}
\end{figure}

We propose the novel \glsfirst{AMT} approach, a system-agnostic method for tracking blinking light sources attached to multi-robot team members.
By using motion models and constraints of cooperating aerial robots, the \gls{AMT} approach solves the tracking problem by fusing the past motion of blinking light sources to estimate the next expected location of team members in the image.
This proposed approach enables vision-based relative localization and \gls{OCC} using blinking light sources attached to fast-moving \glspl{UAV} (Fig.~\ref{fig:intro:overview}).
The source code\footnote{\url{https://github.com/TimLakemann/ami.git}} and a demonstration video\footnote{\url{https://mrs.fel.cvut.cz/towards-agile-swarming-in-real-world}} can be found online.

\section{State of the Art and Contribution}\label{sec:sota}
Active blinking markers attached to the \gls{UAV} frame enhance reliability and enable unique identification, while also reducing the computational cost of object detection algorithms.
In~\cite{whiteOpticalSpatialLocalization2019}, White \emph{et al.} tracked a static \gls{LED} ring using spatial-temporal difference images, with the ring blinking at half the frame rate of the camera. 
This work was extended in~\cite{whiteRobustOpticalSpatial2019} by attaching the \gls{LED} ring to a \gls{UAV} and tracking it with a stationary camera.
However, using difference images restricts the blinking frequency for all team members to half the frame rate of the camera, making it difficult to distinguish between the transmitting \glspl{UAV} (\glsentryshortpl{TX}).

Breitenmoser \emph{et al.} developed a mutual relative localization system using active and passive markers in~\cite{breitenmoserMonocularVisionbasedSystem2011}, achieving centimeter-level accuracy in indoor experiments. 
However, they observed that different marker frequencies could enhance robustness in differentiating robot targets. 
Additionally, the system was not tested for cross-talk detection when robots were close.

In~\cite{stuckeyOpticalSpatialLocalization2021, stuckeySpatialLocalizationAttitude2022,censiLowlatencyLocalizationActive2013,ebmerRealtime6DoFPose2024}, the authors tracked a \gls{UAV} with blinking markers using a \gls{DVS}, which requires different tracking approaches compared to \gls{CMOS} cameras.
Cenci \emph{et al.} extracted the individual frequencies associated with \gls{IR}-markers attached to a \gls{UAV} in an indoor environment~\cite{censiLowlatencyLocalizationActive2013}.
In~\cite{ebmerRealtime6DoFPose2024}, a \gls{DVS} was used to track a marker board with four blinking \glspl{LED}. 
They tested the system in indoor and outdoor environments with a working distance of up to 10 meters.
However, their system does not address occlusions or situations where the \glspl{LED} appear close together in the image. In addition, \glspl{DVS} usually have a limited \gls{FOV} and are more expensive than \gls{CMOS} cameras~\cite{walterFastMutualRelative2018a}. 
To our knowledge, systems using \glspl{DVS} do not handle tracking when both the \gls{RX} and \glsentryshort{TX} are moving, making them impractical for mutual relative localization and, consequently, for multi-robot systems.

In~\cite{gorseOpticalCameraCommunication2018}, the authors changed the light intensity of \gls{IR}-\glspl{LED} to represent the two different bit states.
Changes in light intensity are effective in indoor environments.
However, it causes ambiguities in outdoor environments due to higher background brightness and noise caused by sunlight.

In~\cite{diasOnboardVisionbased3D2016}, the authors used a combination of three blue non-blinking markers for relative localization and one red blinking marker for identification attached to the frame of a \gls{UAV} for indoor relative localization.
This combination enables continuous tracking by the non-blinking markers and unique identification by the blinking markers.
However, non-blinking markers reduce transmission bandwidth and reliability. At the same time, reliance on the visible spectrum increases dependence on lighting conditions and limits the operational range of \glspl{UAV} in outdoor environments.

\subsection{The UVDAR system}\label{sec:uvdar}
\begin{figure}[t]
  \centering
    \includegraphics[width=0.6\textwidth]{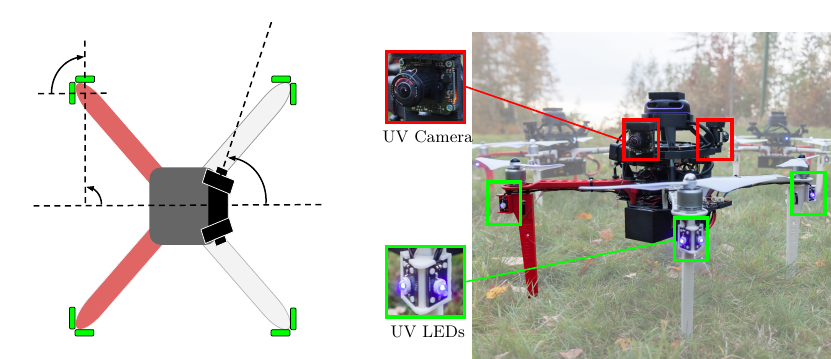}
      \caption{\gls{UAV} equipped with the UVDAR system~\cite{liceaOpticalCommunicationbasedIdentification2023}. }
      \label{fig:uav_platform}%
\end{figure}

The \glsfirst{UVDAR} system provides relative localization and communication for \glspl{UAV} operating in both indoor and outdoor environments~\cite{walterUVDARSystemVisual2019, walterFastMutualRelative2018a, walterMutualLocalizationUAVs2018, petracekBioinspiredCompactSwarms2021,horynaUVDARCOMUVBasedRelative2022}.
The system uses \gls{UV}-\glspl{LED} on the arms of a target \gls{UAV} for data transmission in combination with calibrated grayscale cameras with \gls{UV} band-pass filters (Fig. \ref{fig:uav_platform}) attached to observer \glspl{UAV}~\cite{liceaOpticalCommunicationbasedIdentification2023}.
The band-pass filters remove the most visible light, making the blinking markers appear as bright white spots in the camera image of the observer.
These are then processed using a \gls{FAST}-like procedure and non-maxima suppression to extract the center pixel of each marker~\cite{walterFastMutualRelative2018a,walterMutualLocalizationUAVs2018}. 

Multiple \gls{UV}-\glspl{LED} on a single multirotor's arm emit an identical sequence, recognized as a blinking marker (green box in Fig. \ref{fig:uav_platform}).
These binary sequences are stored in a dictionary \D{} that contains \gls{LED}-IDs associated with each sequence~\cite{liceaOpticalCommunicationbasedIdentification2023}.
In our previous work~\cite{walterMutualLocalizationUAVs2018}, we solved the tracking problem of the blinking markers by using the Hough Transform for 3D line extraction.
The search for maxima in the Hough Space introduces high computational load and memory usage. 
Additionally, the line approximation of the movement of the marker in the image is insufficient for fast and agile maneuvers of the multi-robot system.

\subsection{Contribution}\label{subsec:contribution}
This work is motivated by the need for robust mutual visual tracking in agile multi-robot systems operating in real-world indoor and outdoor environments, particularly within tightly cooperating swarms.
\glspl{UAV} equipped with active blinking markers enable \gls{OCC} and relative localization offering substantial benefits for agile multi-robot systems.
However, effective deployment in agile multi-robot scenarios requires solving three challenges simultaneously: associating multiple anonymous blinking light sources, tracking them in real time, and extracting their individual blinking sequences.
To the best of our knowledge, no existing approach addresses all three challenges concurrently in agile, decentralized multi-robot systems employing active blinking markers.

In this work, the approach was integrated with the \gls{UVDAR} system~\cite{walterFastMutualRelative2018a}, but it can also be applied to any multi-robot framework using active blinking markers.
We validated its effectiveness through outdoor experiments, demonstrating substantial performance improvements over the state-of-the-art method.
The main contributions of this work are the following:
\begin{enumerate}
    \item We present a vision-only technology for tracking blinking markers across consecutive images, enabling onboard mutual relative localization and reliable \gls{OCC} in fast, agile multi-robot aerial systems.
    \item We use uncertainty estimates from weighted polynomial regression to define a search window for future marker appearances, enhancing tracking accuracy under dynamic conditions.
    \item We propose a recovery mechanism that re-tracks blinking markers after tracking failures, significantly increasing reliability for the real-world deployment of closely cooperating teams. 
\end{enumerate}

\section{\glsentrylong{AMT} (\glsentryshort{AMT})}\label{sec:amt}
The proposed \glsfirst{AMT} approach enables the tracking of multiple moving blinking light sources across consecutive image frames, allowing the extraction of individual blinking sequences, which is essential for agile multi-\gls{UAV} systems.
The foundation of the \gls{AMT} approach is the dynamic buffer, \B{}, which stores correspondences of past pixel positions of detected bright points from previous camera frames.
In the following, a distinction is made between \ts{}, \emph{p-states} and \emph{image-points}:
\begin{itemize}
  \item A \ts{} is represented as a row in \B{}, containing information about extracted bright points associated with the same marker emitted by transmitting team members. 
  Each \ts{} consists of multiple \emph{p-states}.
  \item A \ps{} represents the state of a marker tracked by a \ts{} at a specific timestamp. 
  It includes the image capture time, the pixel coordinates and its \emph{state}, which can be either `1' (marker \enquote{on}) or `0' (marker \enquote{off}).
  \item An \emph{image-point} is a binarized bright-white point (see Sec.~\ref{sec:uvdar}) extracted from the latest camera frame, which has not yet been associated with any \ts{} in \B{}.
\end{itemize}
The \emph{image-points} at the latest image capture timestamp $t$ can be denoted in the set:
\begin{align}
  \mathcal{P}_t = \{p_{t,1},\hdots, p_{t,k}, \hdots,p_{t,m}\},
\end{align}
where each $p_{t,k}$ denotes the pixel position of the \emph{image-point}.
The different stages of correspondence searches between \ts{} in \B{} and \emph{image-points} in $\mathcal{P}_t$ are illustrated in Fig.~\ref{fig:amt_overview}. 
If an \emph{image-point} is successfully associated with a \ts{}, it is converted to a \ps{} upon insertion into \B{} (Fig.~\ref{figamtbuffer}).
\begin{figure}
  \centering
  \vspace{-8pt}
  \scalebox{0.5}{\input{algorithm_overview.tex}}
  \caption{Overview of the \gls{AMT} approach: Two correspondence searches -- \emph{Local Search} and \emph{Extended Search} -- associate \emph{image-points} ($\mathcal{P}_t$) with \emph{p-states} stored in \ts{} within \B{}. Finally, the \emph{Verification} method evaluates whether the \ts{} in \B{} are valid.}
    \label{fig:amt_overview}
\end{figure}
The functionality of \B{} and the concept of a \ts{} are discussed in further detail in Sec.~\ref{subsecamtbuffer}.
The \gls{AMT} approach is divided into three parts:
\begin{enumerate}
  \item \emph{Local Search} (Sec.~\ref{subsecamtlocal_search}): uses the expected maximal speed of a marker in the image to approximate their next appearance.
  \item \emph{Extended Search} (Sec.~\ref{subsecamtextended_search}): processes all \ts{} for which the \emph{Local Search} has failed.
  It predicts the next occurrence of a blinking marker based on its past image coordinates.
  \item \emph{Verification} (Sec.~\ref{subsecamtverification}): ensures that \B{} stays within memory bounds and optimizes computational efficiency.
\end{enumerate}

\begin{figure}
  \centering
  \vspace{-8pt}
  \scalebox{0.5}{\input{buffer.tex}}
    \caption{\B{} contains multiple \ts{} ($S_1$, $\hdots,S_{n}$), each with multiple \emph{p-states}. Black circles indicate \enquote{on}-state; white circles indicate \enquote{off}-state of blinking marker. Red rectangle: new \ps{} inserted into $S_i$.}
  \label{figamtbuffer}
\end{figure}

\subsection{Dynamic Buffer \B{}}\label{subsecamtbuffer}

The buffer \B{} serves as the foundation of the tracking method, storing \emph{p-states} extracted from consecutive camera images.
Each \ts{} is represented as a row in \B{}, which contains a sequence of \emph{p-states} associated with a single moving blinking marker.
The \emph{p-states} corresponding to the same timestamp are aligned in columns (Fig. \ref{figamtbuffer}).
\begin{figure}[b]
  \centering
  \vspace{-4pt}
  \scalebox{0.521}{\input{buffer_detailed.tex}}
  \vspace{-3mm}
  \caption{\ts{} containing multiple \emph{p-states}. Red rectangle: new \ps{} inserted at timestamp $t$. Green rectangle: Maximal length $L_S$ of \ts{}. \emph{Verification} method removes \ps{} surpassing $L_S$.}
  \label{figamtbuffer_detailed}
\end{figure}
A \ts{} in the $i$-th row of \B{} is denoted as $S_i$, and a \ps{} within $S_i$ at timestamp $t-j$ (where $j$ indicates the number of frames before the latest image capture at $t$) is written as $p(S_i)_{t-j}$.
When an \emph{image-point} from $\mathcal{P}_t$ is associated with $S_i$, it is appended as a new \ps{} at the end of $S_i$ (red rectangles in Figs.~\ref{figamtbuffer} and~\ref{figamtbuffer_detailed}).

To manage its dynamic nature, \B{} is constrained by two limits: one limits the maximum number of rows ($m_r$), controlling the number of \ts{} stored in \B{}. 
This prevents memory overflow when the image is excessively noisy (e.g, sun reflections on water surface), which could lead to infeasible associations. 
The second constraint is the column limit ($L_S$), which defines the maximum number of \emph{p-states} a \ts{} can store in \B{}. 
The parameter is primarily influenced by the \emph{Extended Search} (Sec.~\ref{subsecamtextended_search}).
By maintaining these constraints, \B{} remains computationally efficient and robust against tracking failures.

\subsection{Local Search}\label{subsecamtlocal_search}
The \emph{Local Search} matches \ts{} in \B{} with \emph{image-points} in $\mathcal{P}_t$ by considering the maximum expected linear displacement $(\Delta px_{m}=(\Delta x_m,\Delta y_m))$ of a blinking marker between two consecutive frames, where $\Delta x_m$ and $\Delta y_m$ represent the horizontal and vertical displacements in pixel units, respectively.
This search is centered around the last known position of the marker to account for its movement.
The maximal expected displacement between two frames is defined as:
\begin{align}
\Delta px_{m} = 
(\lceil v_{x,\max}/f \rceil, \lceil v_{y,\max}/{f}\rceil)\label{eqamtlocal_search},
\end{align}
where $v_{x,\max}$ and $v_{y,\max}$ denote the maximum horizontal and vertical velocities of a blinking marker in the image (in pixels per second), and $f$ represents the camera frame rate.
These values were experimentally determined by analyzing marker motions across frames under various conditions, including different robot speeds, distances, and relative movements.
For each \ts{} in \B{}, the \emph{Local Search} defines a fixed search area around the last inserted \ps{} using equation~\eqref{eqamtlocal_search}.
The search area for a given \ts{}, $S_i$, is given by:
\begin{align}
  p(S_i)_{t-1}\pm \Delta px_{m}.
\end{align} 
If an \emph{image-point}, $p_{t,k}$ in $\mathcal{P}_t$ falls within this search area, it is appended to $S_i$ as a new \ps{}.

This fixed search area is effective when the relative movement between the \gls{RX} and \glsentryshort{TX} results in a small marker displacement in the image.
However, it may not be sufficient when a \gls{UAV} performs agile maneuvers, causing a larger marker displacement.
To handle cases where the \emph{Local Search} fails, all unmatched \emph{image-points} from $\mathcal{P}_t$ are collected in the subset $\mathcal{P}_t^*$. 
Similarly, all \ts{} in \B{} that fail to find a corresponding \emph{image-point} are stored in the subset \BS{}.
These subsets, $\mathcal{P}_t^*$ and \BS{}, are then processed in the subsequent \emph{Extended Search}. 

\subsection{Extended Search}\label{subsecamtextended_search}
The \emph{Extended Search} performs a correspondence search between the \ts{} in \BS{} and the \emph{image-points} in $\mathcal{P}_t^*$.
Within \BS{}, only the \enquote*{$1$}-\emph{p-states} (blinking marker \enquote{on}; Fig.~\ref{figamtextended_search}) are selected, forming a discontinuous and shorter \ts{}, denoted as $S_i^+$. 
The \enquote*{1}-\emph{p-states} provide more precise pixel coordinates than \enquote*{$0$}-\emph{p-states}, where the blinking marker is \enquote{off}.
Each \ps{}, $p(S_i^+)_{t-j}$, where $j$ represents the number of past camera frames relative to the current timestamp $t$, is assigned a weight using an exponential decay function:
\begin{align} 
  w_{t-j} = \exp\left(-\lambda(t-j)\right), \label{amt:eq:weight}
\end{align} 
where $\lambda$ controls the decay rate~\cite{serwayModernPhysics2005}. 
The weights are normalized and stored in a diagonal matrix $\mathbold{W}$. 
To predict the next pixel location of a marker at timestamp $t$, a weighted polynomial regression is applied on each \ts{} in \BS{}(Fig.~\ref{figamtextended_search_2}).
In the following, the weighted polynomial regression is derived for the $x$-pixel, which is equivalent for the $y$-pixel.
The regression coefficients $\hat{\mathbold{\beta}}$ are estimated via the weighted normal equation:
\begin{align} 
  (\mathbold{X}^T\mathbold{WX})\hat{\mathbold{\beta}} = \mathbold{X}^T\mathbold{W} \mathbold{x}, \label{omt:eq:normalFormWeighted} 
\end{align} 
where $\mathbold{X}$ is the Vandermonde matrix containing past timestamps, and $\mathbold{x}$ contains the corresponding observed $x$-coordinates.
This weighted least squares problem is transformed to an ordinary least squares problem using:
\begin{align} 
  \mathbold{X}_w = \mathbold{W}^{\frac{1}{2}}\mathbold{X},\qquad \mathbold{x}_w = \mathbold{W}^{\frac{1}{2}}\mathbold{x}, \label{dependent_x} 
\end{align}
yielding:
\begin{align}
    \mathbold{X}_w^T \mathbold{X}_w\hat{\mathbold{\beta}} = \mathbold{X}^T_w \mathbold{x}_w.\label{omt:eq:weightedNormalLSQ_mat}
\end{align}
\begin{figure}[t]
  \centering
  \vspace{-8pt}
  \subfloat[]{\scalebox{0.6}{\input{extended_search.tex}}%
  \label{figamtextended_search}}
  \hfil
  \subfloat[]{\scalebox{0.2}{\input{extended_search_2.tex}}%
  \label{figamtextended_search_2}}
  \caption{(a) The \emph{Extended Search} selects all \emph{p-states} of a \ts{} with value `1'. (b) Past images (blue) with pixel locations of $S_i^+$ and polynomial regression (red) with its search window in the image at timestamp $t$.}
  \label{fig_sim}
  \vspace{-15pt}
\end{figure}

Since direct inversion of $\mathbold{X}^T_w\mathbold{X}_w$ is numerically unstable~\cite{ComputationalAlgorithmsFitting2003}, the problem is reformulated using \emph{QR}-Decomposition using \emph{Householder}-reflections resulting in:
\begin{align} 
  \mathbold{R}\hat{\mathbold{\beta}} = \mathbold{Q}^T \mathbold{x}_w \label{amt:eq:finalSol} 
\end{align}
which is then solved using backward substitution~\cite{golubMatrixComputations1996}.

For each \ts{} in \BS{}, an individual search window is constructed based on the uncertainty in computing the regression coefficients, referred to as the prediction interval for a new response~\cite{rossChapterRegression2014}. 
This interval assumes that the estimation error follows a normal distribution and is derived from the standard error between the unknown value $x_t$ and its estimate $\hat{x}_t$:
\begin{align} 
  SE_{x_{t}} = \sqrt{\hat{\sigma}_w^2\left(1+\frac{1}{L_i^+}+\frac{(t-\bar{t}_w)^2}{\sum_{k=1}^{L^+_i}{(t_k-\bar{t}_w)^2}}\right)},\label{amt:se_w}
\end{align}
where $\bar{t}_w$ is the weighted mean of the timestamps in $S_i^+$, $L_i^+$ is the length of $S_i^+$, and $\hat{\sigma}_w^2$ is the weighted unbiased estimator for the error variance ($\sigma_w^2$), given by:
\begin{align}
    \hat{\sigma}_w^2 = \frac{\sum_{k=1}^{L^+_i}{w_k(x_k - \hat{x}_k)^2}}{L^+_i-(d+1)}\label{amt:eq:w_sigma_square}.
\end{align}
Using equation~\eqref{amt:se_w} and the Student's \texttt{t}-distribution (\texttt{t}$_{1-\frac{\alpha}{2},\nu}$), the search window around $\hat{p}{(S_i^+)_t}$ for the $x$-pixel at timestamp $t$ is defined by:
\begin{align}
  \hat{x}_t \pm (\text{\texttt{t}}_{1-\frac{\alpha}{2},\nu})\sqrt{\hat{\sigma}_w^2\left(1+\frac{1}{L_i^+}+\frac{(t-\bar{t}_w)^2}{\sum_{k=1}^{L_i^+}{(t_k-\bar{t}_w)^2}}\right)},\label{eqamtextended_search}
\end{align}
where $\alpha$ is the significance level determining the critical value from the Student's \texttt{t}-distribution, and $\nu$ is its degree of freedom, defined as:
\begin{align}
  \nu = L_i^+ - (d + 1),
\end{align}
with $d$ representing the polynomial degree used in the regression.

The maximum permissible length of a \ts{} in \B{}, denoted by $L_S$, depends on the parameters of the \emph{Extended Search}. 
The choice of polynomial degree $d$, which is essential to approximate the past movement of a blinking marker, depends on the number of past \emph{p-states} stored in a \ts{}.
Consequently, a larger $L_S$ results in a longer motion history, necessitating a higher polynomial degree and vice versa.
This relationship can be expressed by:
\begin{align}
L_S = \eta d\label{eqamtseq_length},
\end{align}
where $\eta \in \mathbb{N}^+$ is an experimentally determined parameter.

Similarly to the \emph{Local Search}, if an \emph{image-point} from $\mathcal{P}_t^*$ lies in the search window of a \ts{}, defined by equation \eqref{eqamtextended_search}, it is appended to the end of that \ts{}.
The \emph{image-points} from $\mathcal{P}_t^*$ that remain unmatched after the \emph{Extended Search} are stored in a new set, denoted $\mathcal{P}_t^\Gamma$. 
Similarly, \emph{t-series} from \B{}$^*$ without new insertions from $\mathcal{P}_t^*$ are stored in a new set, \B{}$^\Gamma$.

\subsection{Verification}\label{subsecamtverification}
The \emph{Verification} method performs multiple tasks.
If \B{}$^\Gamma$ contains \ts{}, both correspondence searches failed to find new associations for the stored \ts{}.
\begin{figure}[!b]
  \centering
  \vspace{-5pt}
  \scalebox{0.45}{\input{verification_buffer.tex}}
  \caption{$S_1$ and $S_n$ in gray: successful correspondence searches by the \emph{Local Search} or \emph{Extended Search}. Red arrow: duplication of $p(S_i)_{t-1}$ to `$0$'-\ps{} at timestamp $t$. Green rectangle: initialization of a new \ts{} at the end of \B{}.} 
  \label{fig:verification}
\end{figure}
For a blinking marker represented by the \ts{} in \B{}$^\Gamma$, it is expected that the marker is either \enquote{off} in the image frame at timestamp $t$ or otherwise not visible.
Consequently, for all \ts{} in \B{}$^\Gamma$, the pixel coordinates of the \emph{p-states} at timestamp $t-1$ are duplicated and inserted as new \enquote*{$0$}-\emph{p-states}.
Since the \emph{Extended Search} ignores \enquote*{$0$}-\ps{}, the previous motion is not considered when inserting the \enquote*{$0$}-\ps{}.
In Fig.~\ref{fig:verification}, a red arrow highlights this process for $S_i$.
This could result in continuously inserting \emph{p-states} representing \enquote*{$0$} bits into \ts{} that lack new \emph{image-point} associations.
This scenario is likely to occur when a blinking marker exits the \gls{FOV} of the camera or if it becomes occluded, both of which would result in a \ts{} containing only `0'-\emph{p-states}.
In the \emph{Verification} method, the validity of each \ts{} in \B{} is confirmed by the condition:
\begin{align}
  \sum_{j=t-(e+b_{m,0})}^{t} p_\texttt{s}(S_i)_j \geq e+b_{m,0} \label{eqamtverification} \hspace{1cm} \{i\in \mathcal{B}\},
\end{align}
where $p_\texttt{s}(S_i)_j$ represents the \emph{state} of $p(S_i)$ at timestamp $j$, $b_{m,0}$ denotes the allowed maximum bits of consecutive zeros in a \ts{}, and $e$ corresponds to the expected bit error rate per sequence transmission.
If condition~\eqref{eqamtverification} is violated for a \ts{} due to continuous \enquote*{$0$}-\ps{} insertions, the \emph{Verification} method removes the \ts{} from \B{}.
To re-track a blinking marker after a tracking failure caused by situations such as obstruction by an object, the value of $b_{m,0}$ should be set sufficiently high to keep the invalid \emph{t-series} in the buffer.
This allows for re-tracking the blinking marker by the \emph{Extended Search}.
However, increasing the value of $e$, the overall computational cost and memory usage increases, since invalid \emph{t-series} are kept longer in \B{} until condition \eqref{eqamtverification} is violated.
Additionally, the \emph{Verification} method removes \emph{p-states} in a \ts{} that exceed the maximum allowed number of columns, $L_S$, thereby preventing memory overflow (Fig.~\ref{figamtbuffer_detailed}).
The maximum number of rows, and consequently the maximum number of \ts{} in \B{}, is defined by $m_{r}$. 
Each \emph{image-point} in $\mathcal{P}_t^\Gamma$ is inserted as a new \ts{} into \B{} if the row limit ($m_r$) is not exceeded.
This increases the number of rows by the length of $\mathcal{P}_t^\Gamma$, enabling the algorithm to track newly appearing blinking markers in the \gls{FOV} of the camera.
Fig.~\ref{fig:verification} shows this process for a single \emph{image-point} by the green rectangle.

By this stage, all \emph{image-points} in $\mathcal{P}_t$ have been inserted into \B{}, either through the \emph{Local Search}, \emph{Extended Search}, or \emph{Verification} method. 
As a final step, it is verified whether a \ts{} in \B{} matches a sequence in \D{}.
If a match is found, the \ts{} is associated with the corresponding ID of that sequence in \D{}.

\section{Experimental Evaluation}\label{sec:evaluation}
\begin{figure*}[t]
  \centering
  \subfloat[\label{fig:eval:two_tx}]{
    \scalebox{1.1}{\input{outdoor_2.tex}}
  } 
  \subfloat[\label{fig:eval:overview_graph}]{
    \scalebox{0.4}{\input{experiment_setup.tex}}}
    \caption{(a) Two \glsentrylongpl{TX} flying in front of one \glsfirst{RX}. 
    (b) Experimental Setup:
    In experiments 2-6, the right camera (red) of the \gls{RX} points at two \glsentryshortpl{TX} flying parallel to the $y$-axis.
    In Experiment 7, only one \gls{TX} is used to test the algorithm during agile maneuvers.
    Numbers around the \glsentryshortpl{TX}' arms show the \gls{LED}-IDs emitted by the blinking markers.}
    \label{fig:eval:overview}
    \vspace*{-10pt}
\end{figure*}
We evaluated the \gls{AMT} approach in outdoor experiments, comparing it with the state-of-the-art \gls{4DHT} approach~\cite{walterMutualLocalizationUAVs2018} used in the previous version of the \gls{UVDAR} system. 
A quadrotor \gls{UAV} based on the \emph{Holybro X500} platform was used with an \emph{Intel NUC 10 i7FNK} (6 cores, up to \SI{4.7}{\giga\hertz}; details in~\cite{hertMRSDroneModular2023}).
A \gls{RTK} base station was used to obtain ground-truth measurements.
Both onboard and offline executions showed indiscernible performance differences.
Therefore, we re-executed both algorithms on the same computer (\emph{Intel i7-8550U} CPU, \SI{1.8}{\giga\hertz}) and dataset for a fair comparison.
This paper presents the seven most relevant experiments from a set of 14, conducted with two or three \glspl{UAV} (Fig.~\ref{fig:eval:overview}).
Each trajectory in each experiment was flown in a periodic loop, with a minimum duration of \SI{60}{\second}. 
The experiments included the following flight trajectories:

Exp. 1 -- Yaw Rotation: The \gls{RX} rotated around its yaw axis while two \glspl{TX} hovered at approximately \SI{4}{\meter} and \SI{8}{\meter}, resulting in linear horizontal marker motion in the image.

\begin{wrapfigure}{r}{0.25\textwidth}
  \scalebox{0.5}{\input{star_setup.tex}}
  \caption{$y$-$z$ coordinates for \glsentryshort{TX}1 and \glsentryshort{TX}2 during \enquote{star}-shaped trajectory in Exp. 6.} 
  \label{fig:eval:star}
  \vspace*{-14pt}
\end{wrapfigure}
Exp. 2 -- Parallel Motion with Rotation: All \glspl{UAV} followed an \SI{8}{\meter} linear trajectory, with the \glsentryshortpl{TX} rotating by \SI{180}{\degree} and the \gls{RX} rotating between \SI{0}{\degree} and \SI{90}{\degree}, introducing additional horizontal motion and marker crosstalk.

Exp. 3 -- Linear \gls{RX} Motion: The \gls{RX} moved \SI{4}{\meter} along the $x$-axis (see Fig.~\ref{fig:eval:overview_graph}), causing vertical marker motion and enabling evaluation of the algorithms under abrupt deceleration and acceleration.

Exp. 4 -- Perpendicular Motion with Occlusion: The \glsentryshortpl{TX} moved \SI{8}{\meter} perpendicular to the camera axis, reaching maximum speeds of \SI{1.19}{\meter\per\second} and \SI{1.54}{\meter\per\second}, respectively. 
Occlusions occurred at the intersections of the trajectory within the image.

Exp. 5 -- Circular Motion: The \glsentryshortpl{TX} followed circular trajectories with radii of \SI{1}{\meter} (maximum velocity: \SI{0.6}{\meter\per\second}) and \SI{1.5}{\meter} (maximum velocity: \SI{1.44}{\meter\per\second}), introducing curved trajectories and occlusions.

Exp. 6 -- Star Trajectory: The \glsentryshortpl{TX} followed a \enquote{star}-shaped trajectory (Fig.~\ref{fig:eval:star}) reaching maximum velocities of \SI{2.22}{\meter\per\second} and \SI{1.63}{\meter\per\second}, respectively.

\begingroup
\setlength{\tabcolsep}{5pt}  
\renewcommand{\arraystretch}{1.} 
\begin{table}[]
  \centering
  \begin{tabular}{c c c c}
    \toprule
    \textbf{Exp.} & \textbf{Duration [s]} & \textbf{Weather} & \textbf{Location} \\
    \midrule
    1 & 118 & Twilight, scattered clouds & \multirow{6}{*}{$49.362^\circ\text{N}; 14.261^\circ\text{E}$} \\
    2 & 70 & Twilight, scattered clouds & \\
    3 & 72 & Twilight, scattered clouds & \\
    4 & 70 & clear & \\
    5 & 60 & clear & \\
    6 & 60 & clear & \\
    \midrule
    7 & 162 & scattered clouds & $50.112^\circ\text{N}; 14.418^\circ\text{E}$ \\
    \bottomrule
  \end{tabular}
  \caption{Environmental conditions, including duration, weather, and location for each experiment.}
  \label{tab:eval:envionment_conditions}
\end{table}
\endgroup
\begingroup
\setlength{\tabcolsep}{3pt}
\renewcommand{\arraystretch}{1.05} 
\begin{table}[b]
  \centering
  \small
  \begin{tabular}{c r r r r r r r r r r r}
    \toprule
    \textbf{Exp.} & \multicolumn{1}{c}{$\boldsymbol{f}$} & \multicolumn{1}{c}{$\boldsymbol{b_{m,0}}$} & \multicolumn{1}{c}{$\boldsymbol{e}$} & \multicolumn{1}{c}{$\boldsymbol{\Delta px_{m}}$} & \multicolumn{1}{c}{$\boldsymbol{\alpha}$} & \multicolumn{1}{c}{$\boldsymbol{L_\mathcal{D}}$} & \multicolumn{1}{c}{$\boldsymbol{L_S}$} & \multicolumn{1}{c}{$\boldsymbol{d}$} & \multicolumn{1}{c}{$\boldsymbol{\eta}$} & \multicolumn{1}{c}{$\boldsymbol{\lambda}$} & \multicolumn{1}{c}{$\boldsymbol{m_r}$}\\
    & [fps] & [bits] & [bits] & [px, px] & [\%] & [bits] & [bits] & & &  \\
    \midrule
    1+2 & 60 & 10 & 0 & [0, 7] & 80.0 & 18 & 360 & 1 & 360 & 1.0 & 500 \\
    3--6 & 60 & 10 & 0 & [3, 3] & 95.0 & 18 & 360 & 4 & 90 & 0.1  & 500\\
    7 & 60 & 10 & 0 & [6, 6] & 95.0 & 18 & 360 & 3 & 120 & 0.1  & 500\\
    \bottomrule
  \end{tabular}
  \caption{Parameter settings for AMT experiments.}
  \label{tab:eval:params}
\end{table}
\endgroup
Exp. 7 -- Agile Motion Trajectories: A single \glsentryshort{TX} followed linear (7.1), circular (7.2), and \enquote{star}-shaped (7.3) trajectories at two different speeds, reaching a maximum velocity of \SI{5.43}{\meter\per\second} at a Euclidean distance of \SI{4}{\meter} from the \glsentryshort{RX}.

\begin{figure*}[!t]
  \centering
  \includegraphics[width=1.\textwidth]{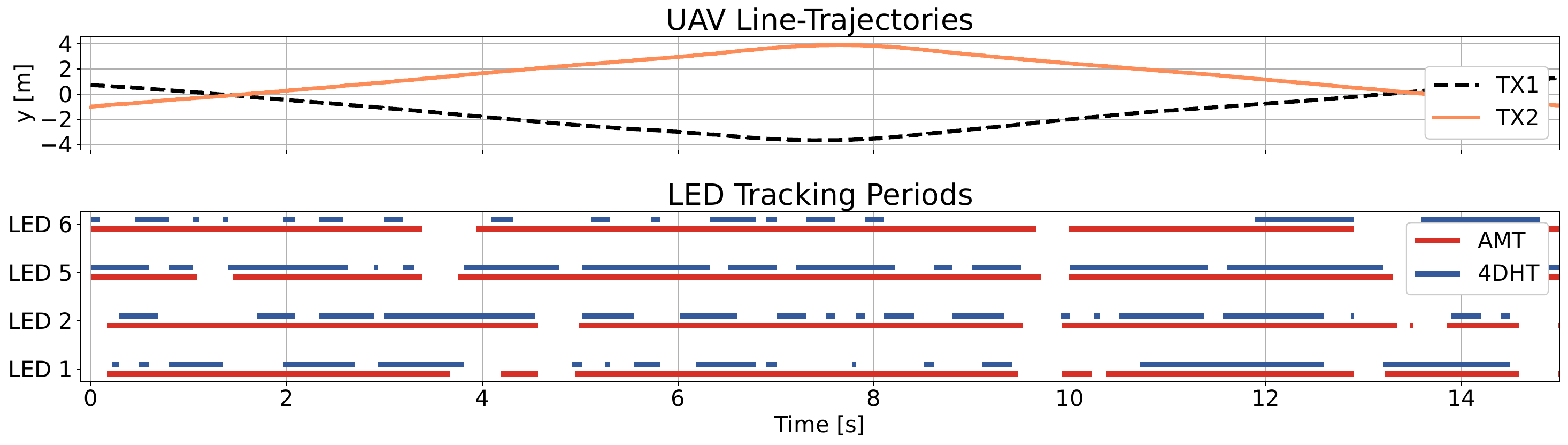}
  \vspace{-16pt}
  \caption{Experiment 4: Extracted \gls{LED}-IDs with y-coordinates for \gls{TX}1 and \gls{TX}2 during linear trajectory.} 
  \label{fig:eval:exp_4}
\end{figure*}

\begin{figure*}[!t]
  \centering
  \includegraphics[width=1.\textwidth]{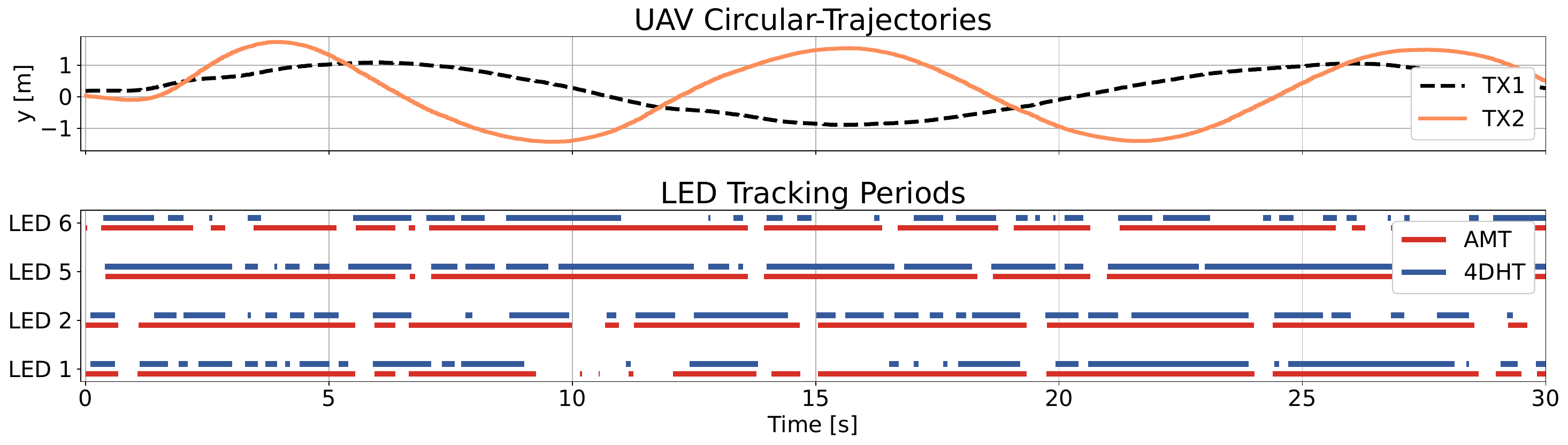}
  \vspace{-16pt}
  \caption{Experiment 5: Extracted \gls{LED}-IDs with y-coordinates for \gls{TX}1 and \gls{TX}2 during circular trajectory.} 
  \label{fig:eval:exp_5}
\end{figure*}

\begin{figure*}[!t]
  \centering
  \includegraphics[width=1.\textwidth]{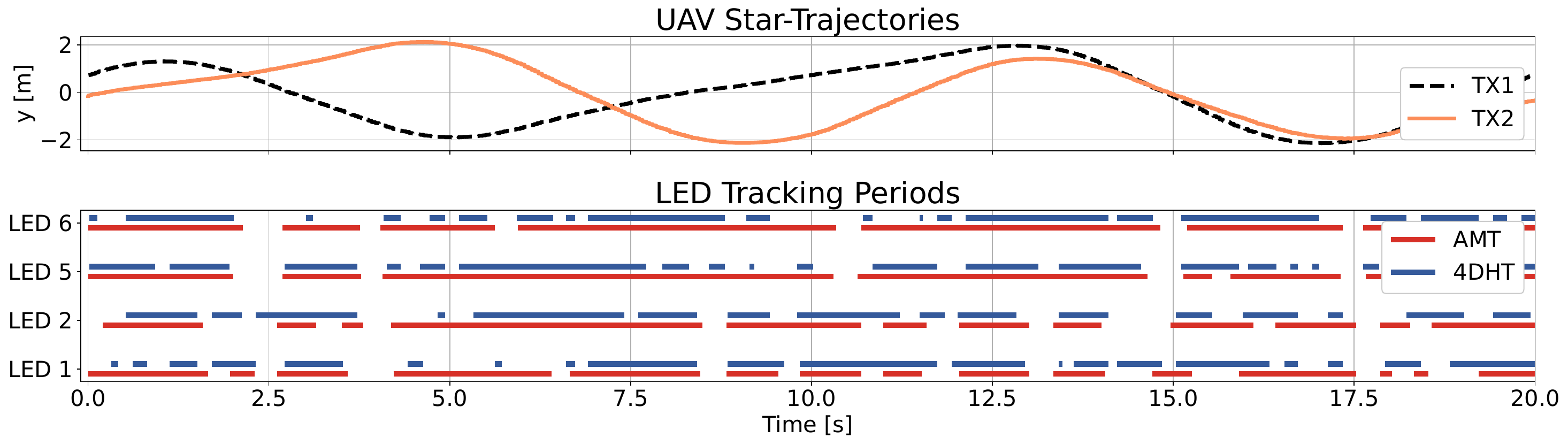}
  \vspace{-16pt}
  \caption{Experiment 6: Extracted \gls{LED}-IDs with y-coordinates for \gls{TX}1 and \gls{TX}2 during \enquote{star}-shaped trajectory.} 
  \label{fig:eval:exp_6}
\end{figure*}

\begin{figure*}[!t]
  \centering
    \includegraphics[width=\textwidth]{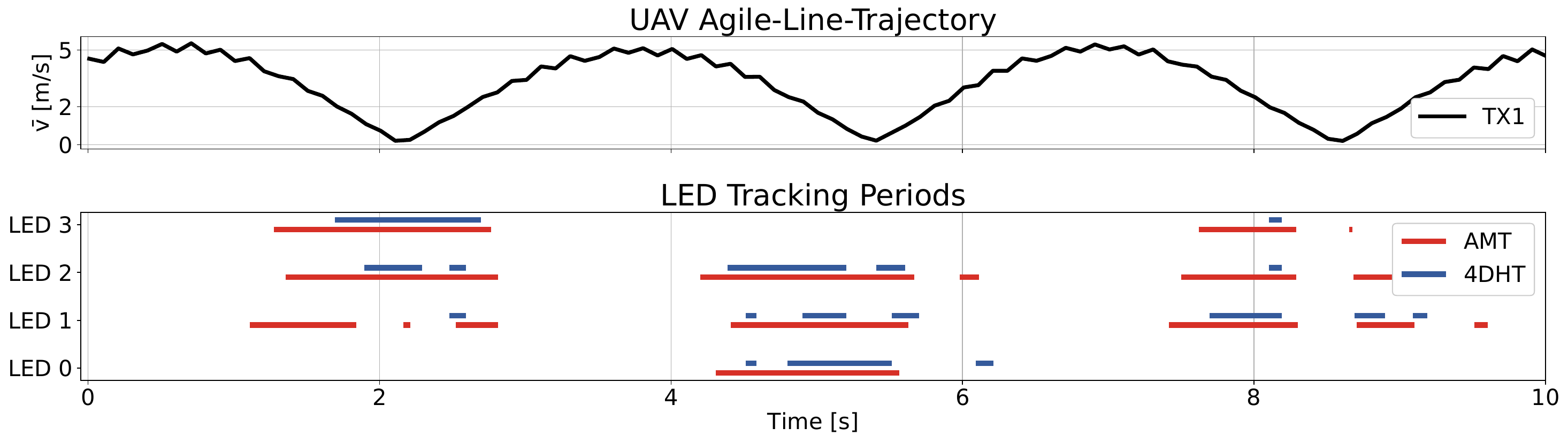}
    \vspace{-16pt}
    \caption{Experiment 7.1: Extracted \gls{LED}-IDs with the relative velocity of \gls{TX}1 during the fastest linear trajectory.}
  \label{fig:eval:exp_7}
\end{figure*}

Tab.~\ref{tab:eval:envionment_conditions} provides an overview of the experimental setup and conditions, including duration, weather, and location.
The parameters selected for the \gls{AMT} approach are shown in Tab.~\ref{tab:eval:params}.
These were chosen based on expected flight dynamics, trajectory patterns, and previous simulation tests.
One of the most critical parameters is the maximum expected movement of a marker between two camera frames ($\Delta px_m$), which determines the size of the search window for the \emph{Local Search}.
For example, in experiments 1 and 2, which involved horizontal linear motions due to rotation around the z-axis of the \gls{UAV}, the \emph{Local Search} area was restricted to a horizontal line, and the polynomial degree for the \emph{Extended Search} was set to $1$.
Increasing the \emph{Local Search} area enables tracking of more agile maneuvers but also increases the risk of failure when multiple \glspl{UAV} are in close proximity within the image. 
To address this, a set of universal parameters was designed for experiments 3 to 6, to balance tracking agility in both vertical and horizontal directions and robustness in scenarios with \glspl{UAV} close together.
These parameters offered the best trade-off: the \emph{Local Search} area was small enough to differentiate the occluding \glspl{UAV} while still allowing the agility required for swarming applications.
Figs.~\ref{fig:eval:exp_4} --~\ref{fig:eval:exp_7} illustrate the \glspl{UAV} trajectories for experiments 4--7, along with the marker periods extracted by each algorithm.
To enhance readability and facilitate comparison, excerpts from the experiments are shown.
Figures~\ref{fig:eval:exp_4}--\ref{fig:eval:exp_6} highlight markers 1, 2, 5, and 6, as these were oriented towards the \gls{RX}.
\begingroup
\begin{table}[t]
  \centering
  \setlength{\tabcolsep}{2.7pt}
  \begin{tabular}{c | l r r r r r r r r | r r}
  \toprule
  \textbf{Exp} & \multicolumn{1}{c}{\textbf{ID:}} & \multicolumn{1}{c}{\textbf{0}} & \multicolumn{1}{c}{\textbf{1}} & \multicolumn{1}{c}{\textbf{2}} & \multicolumn{1}{c}{\textbf{3}} & \multicolumn{1}{c}{\textbf{4}} & \multicolumn{1}{c}{\textbf{5}} & \multicolumn{1}{c}{\textbf{6}} & \multicolumn{1}{c|}{\textbf{7}} & \multicolumn{1}{c}{\textbf{\emph{p}-value}} & \multicolumn{1}{c}{\textbf{\SI{95}{\percent} CI}} \\\midrule
  \multirow{2}{*}{1} &  \glsentryshort{AMT}   &  0.0  &  21.7  &  14.3  &  6.0  &  0.0  &  18.7  &  17.3  &  0.0 & \multirow{2}{*}{ 0.005 } &  \multirow{2}{*}{[6.63, 19.02]}  \\  
  &  \glsentryshort{4DHT}   &  0.0  &  3.4  &  2.2  &  1.2  &  0.0  &  3.7  &  3.4  &  0.0 & &   \\\midrule
  \multirow{2}{*}{2} &  \glsentryshort{AMT}   &  26.9  &  33.8  &  22.5  &  13.0  &  28.2  &  25.5  &  5.8  &  8.7 & \multirow{2}{*}{ 0.001 } &  \multirow{2}{*}{[9.66, 23.96]}  \\  
  &  \glsentryshort{4DHT}   &  4.8  &  4.9  &  3.4  &  2.2  &  6.1  &  5.3  &  1.6  &  1.8  & & \\\midrule
  \multirow{2}{*}{3} &  \glsentryshort{AMT}   &  0.0  &  36.6  &  33.4  &  2.1  &  0.0  &  25.9  &  13.5  &  0.0 & \multirow{2}{*}{ 0.034 } &  \multirow{2}{*}{[1.81, 30.35]}  \\  
  &  \glsentryshort{4DHT}   &  0.0  &  5.1  &  4.1  &  0.1  &  0.0  &  3.8  &  1.8  &  0.0 & & \\\midrule
  \multirow{2}{*}{Occl.} &  \glsentryshort{AMT} &  23.4  &  47.7  &  49.8  &  10.7  &  0.6  &  50.9  &  51.7  &  2.7 & \multirow{2}{*}{ 0.008 } &  \multirow{2}{*}{[9.36, 41.86]}  \\  
  &  \glsentryshort{AMT} &  3.2  &  6.4  &  7.3  &  1.1  &  0.0  &  7.8  &  6.8  &  0.0 & &\\
  \midrule
  \multirow{2}{*}{7.1} &  \glsentryshort{AMT}   &  9.8  &  25.2  &  26.0  &  10.5  & ---   & ---   & ---   & --- & \multirow{2}{*}{ 0.029 } &  \multirow{2}{*}{[3.06, 28.16]}  \\  
  &  \glsentryshort{4DHT}   &  1.5  &  3.2  &  3.1  &  1.2  & ---   & ---   & ---   & ---  & & \\  
  \midrule
  \multirow{2}{*}{7.2} &  \glsentryshort{AMT}   &  16.3  &  36.0  &  32.3  &  13.3  & ---   & ---   & ---   & --- & \multirow{2}{*}{ 0.02 } &  \multirow{2}{*}{[6.31, 36.03]}  \\  
  & \glsentryshort{4DHT}   &  2.3  &  5.6  &  4.3  &  1.0  & ---   & ---   & ---   & ---  & & \\  
  \midrule
  \multirow{2}{*}{7.3} &  \glsentryshort{AMT}   &  11.6  &  18.8  &  21.4  &  3.2 & --- & --- & --- & --- & \multirow{2}{*}{ 0.038 } &  \multirow{2}{*}{[1.16, 21.71]}  \\  
  & \glsentryshort{4DHT}   &  1.4  &  3.8  &  3.8  &  0.3 & --- & --- & --- & --- & &  \\
  \bottomrule
  \end{tabular}
  \caption{Statistical analysis of success rates per experiment duration, including \emph{p}-values and \SI{95}{\percent} \glsfirst{CI}.}
  \label{tab:eval:statistics}
  \end{table}
\endgroup
In all experiments, the \gls{AMT} approach consistently outperformed the state-of-the-art \gls{4DHT} approach, achieving longer and more precise marker tracking.
Tab.~\ref{tab:eval:statistics} further highlights this by presenting the success rate per experiment duration, measured per \gls{LED}-ID.
This rate is calculated as the total number of successful \gls{LED}-ID extractions divided by the duration of the experiment in seconds.
To determine whether the success rates per unit time differed significantly between the two algorithms, we first tested the normality of differences using the Shapiro--Wilk test.
Based on the result, we applied either a paired-\texttt{t}-test (for normally distributed differences) or a Wilcoxon signed-rank test (for non-normal differences).
The resulting \emph{p}-value and \SI{95}{\percent} \gls{CI} indicate the statistical significance of the difference between the \gls{AMT} and \gls{4DHT} algorithms.
To further evaluate performance in challenging scenarios, we analyzed occlusions and marker proximity in experiments 4--6, labeled as \enquote{Occl.} in Tab.~\ref{tab:eval:statistics}.
From these experiments, we extracted a total of \SI{36}{\second} during which the markers were occluded or close to each other in the image.

Both algorithms performed less effectively during agile maneuvers.
Potential reasons for this include the high speed and abrupt stops at the trajectory endpoints (e.g., during linear trajectories), which caused discrepancies between the predicted and actual positions.
Since the \gls{4DHT} approach approximates the movement of markers with straight lines, it fails to capture the dynamic nature of agile maneuvers. 
For the \gls{AMT} approach, a potential reason is that the high speed exceeded the expected maximum marker displacement between consecutive frames, resulting in inaccuracies in polynomial regression predictions.
Furthermore, during highly dynamic motions, the image becomes blurry, which can hinder distinct marker extraction and lead to tracking algorithm failures.
A potential solution is to increase the frame rate and reduce the exposure time.
However, a shorter exposure time reduces the visibility range of the markers, thereby limiting the operational distance between multi-robot team members. 

Among all experiments, experiment 7.3 (agile \enquote{star}-shaped trajectory) had the lowest lower bound of the confidence interval while still remaining entirely positive.
Therefore, with \SI{95}{\percent} confidence, the \gls{AMT} algorithm achieved a success rate between \SI{1.16} and \SI{21.71} units higher than the \gls{4DHT} algorithm, even in the worst-performing case for the \gls{AMT} approach.

\subsection{Computational Complexity and Memory Usage}
Alongside its tracking performance, the \gls{AMT} approach significantly outperforms the state-of-the-art \gls{4DHT} approach in terms of computational efficiency and memory usage.

\begingroup
\begin{table}[t]
  \centering
  \renewcommand{\arraystretch}{1.2} 
  \setlength{\tabcolsep}{2pt}       
  \begin{tabular}{c c | *{6}{c} | c c}
  \toprule
  \textbf{Metric} & \textbf{Method} & \textbf{1 + 2} & \textbf{3} & \textbf{4 - 6} & \textbf{7.1} & \textbf{7.2} & \textbf{7.3} & \textbf{\emph{p}-value} & \textbf{\SI{95}{\percent} \glsentryshort{CI}} \\
  \midrule
  $\overline{RSS}$ & \glsentryshort{AMT} & 141 & 129 & 130 & 157 & 145 & 150 & \multirow{2}{*}{0.031} & \multirow{2}{*}{[-1120, -695]} \\
  $[\text{mB}]$ & \glsentryshort{4DHT} & 866 & 845 & 859 & 1246 & 1238 & 1246 & & \\
  \midrule
  Runtime & \glsentryshort{AMT} & 341 & 309 & 289 & 128 & 170 & 75 & \multirow{2}{*}{$7.0\mathrm{e}{-10}$} & \multirow{2}{*}{[-9002, -8629]} \\
  $[\text{ms/}f]$ & \glsentryshort{4DHT} & 9169 & 9150 & 8833 & 8966 & 9265 & 8823 & & \\
  \bottomrule
  \end{tabular}
  \caption{Average computation time per camera frame and \gls{RSS} per experiment, including \emph{p}-values and \SI{95}{\percent} confidence intervals (\glsfirst{CI}).}
  \label{tab:eval:comparison}
\end{table}
\endgroup

The computational complexity and memory usage of both algorithms depend heavily on predefined parameters.
The \gls{AMT} approach achieves lower complexity primarily by avoiding the exhaustive 4D search required by the \gls{4DHT} approach. 
Although~\cite{walterMutualLocalizationUAVs2018} optimized the \gls{4DHT} approach by reducing the search space to two 2D maxima matrices (of image size), it still involves identifying the maxima for each marker, associating them with the constructed lines, and extracting the blinking sequences.
In contrast, the computational load of the \gls{AMT} approach is primarily determined by the frequency of polynomial regressions performed on \ts{}, as these represent the most computationally expensive operations, which is for the \emph{QR}-Decomposition using \emph{Householder}-reflection $\mathcal{O}(n^3)$ and backward substition $\mathcal{O}(n^2)$, with $n$ being the number of observations used for the regression.
In theory, the runtime complexity of the \gls{AMT} approach could match that of the \gls{4DHT} approach if polynomial regressions were applied to each \ts{} and the number of \ts{} in \B{} exceeded the image resolution.
However, in practice, the number and length of \ts{} in \B{} are much smaller than the image resolution, as the number of extracted markers is minimal compared to the total number of pixels, even in multi-robot swarms with \SI{30} to \SI{50} robots in the camera image. 
Additionally, polynomial regression is not performed on every \ts{}, and those that violate the condition~\eqref{eqamtverification} are removed from \B{}, further reducing computational overhead.

The memory requirements of both approaches differ significantly.
The \gls{AMT} approach maintains a single buffer \B{}, which scales depending on the number of markers in the image.
In contrast, the \gls{4DHT} approach stores two 2D matrices, resulting in a memory complexity of $\mathcal{O}(h \times w \times \phi_d \times \psi_d)$, where $h$ and $w$ describe the height and width of the image, and $\phi_d$ and $\psi_d$ describe the discretization of the two angles used to approximate the lines that represent the movement of the marker~\cite{walterMutualLocalizationUAVs2018}.
This is inefficient because it remains independent of the number of observed markers.
Theoretically, the memory footprint of the \gls{AMT} approach could be artificially increased by modifying the condition~\eqref{eqamtverification} to retain all \ts{} until reaching the maximum length $L_S$.
Furthermore, setting the maximum row length ($m_r$) to twice the image resolution and filling \B{} with \ts{} until the limit is reached could lead to a memory allocation comparable to that of the \gls{4DHT} approach.
However, under reasonable parameters ($L_S < 500$, $m_r < 600$, $b_{m,0} < 50$), experiments confirmed that the \gls{AMT} approach does not exhibit significant memory growth even under extreme conditions.
For instance, when testing with noise images by pointing the camera towards the sun, with small tree branches in the foreground, the approach tracked multiple bright points as \ts{}, yet no measurable memory increase was observed. 
Real-world outdoor experiments further demonstrated that the \gls{AMT} approach required significantly fewer computational and memory resources.
As shown in Tab.~\ref{tab:eval:comparison}, the \gls{AMT} approach required significantly less computation time (\emph{p}-value < 0.05) and memory (\emph{p}-value < 0.01) compared to the state-of-the-art method.

Overall, these results highlight that the \gls{AMT} approach excels in tracking fast, non-linear maneuvers, achieving higher detection frequency while optimizing resource efficiency. 
This makes it particularly well-suited for agile multi-robot teams operating in dynamic environments.

\section{Conclusion}\label{sec:conclusion}
A novel approach, \glsfirst{AMT}, for extracting and tracking moving blinking light sources attached to multi-robot team members was introduced. 
The method was designed to satisfy the requirements and constraints of agile, compact multi-robot systems, tight formation flight, and high-speed multi-\gls{UAV} operations.
We demonstrated its performance in real-world outdoor experiments focusing on agile flight and tested the tracking of multiple \glspl{UAV} on various trajectories, including linear and circular motions, as well as scenarios with mutual occlusions of the \glsentrylongpl{TX}.
The algorithm surpassed the state-of-the-art method in tracking density, accuracy, and efficiency, significantly reducing computational and memory demands.
The higher tracking density supports significantly faster relative motions, up to \SI{5}{\meter\per\second}, making \gls{AMT} ideally suited for agile multi-robot systems.
Thus, we propose this approach as an enabling technology for mutual localization and omnidirectional low-bandwidth visual communication within agile Multi-\gls{UAV} teams.

\bibliographystyle{elsarticle-num}
\bibliography{Manuscript}

\vfill

\end{document}

%% file: tikz_settings.tex
\usetikzlibrary{fadings, fit, backgrounds, calc,arrows, arrows.meta,decorations.markings,spy}

\tikzset{
    cross/.pic = {
    \draw[rotate = 45] (-#1,0) -- (#1,0);
    \draw[rotate = 45] (0,-#1) -- (0, #1);
    }
}

\tikzset{
  camera/.pic = {
    \fill [] (-0.5,-0.5) rectangle (0.5,0.5);
    \draw [] (0.0,0.0) 
    -- (-0.6,1.3) 
    -- (0.6,1.3) 
    -- cycle;
  }
}
\tikzset{
  camera_right/.pic = {
    \fill [red] (-0.5,-0.5) rectangle (0.5,0.5);
    \draw [red] (0.0,0.0) 
    -- (-0.6,1.3) 
    -- (0.6,1.3) 
    -- cycle;
  }
}

\tikzset{
  uav_frame/.pic = {
    \draw [line width=0.6mm] (-1,-1) -- (1,1);
    \draw [line width=0.6mm] (-1,1) -- (1,-1);
    \fill [black] (-0.4,-0.4) rectangle (0.4,0.4);
  }
}

\tikzset{
  observer/.pic = {
    \pic at (0,0) {uav_frame};
    \pic [rotate=70, scale=0.35] at (-0.3,0.35) {camera};
    \pic [rotate=-70, scale=0.35] at (0.4,0.35) {camera_right};
  }
}

\tikzset{
  tx/.pic = {
    \pic at (0,0) {uav_frame};
    \filldraw [red] (-1,-1) circle (3pt);
    \filldraw [red] (-1,1) circle (3pt);
    \filldraw [red] (1,-1) circle (3pt);
    \filldraw [red] (1,1) circle (3pt);
  }
}

%% file: desert.tex
\begin{tikzpicture}[spy using outlines={size=12mm}, rectangle, connect spies]
  \node[anchor=south west,inner sep=0] (a) at (0,0) { 
  \includegraphics[width=0.45\textwidth]{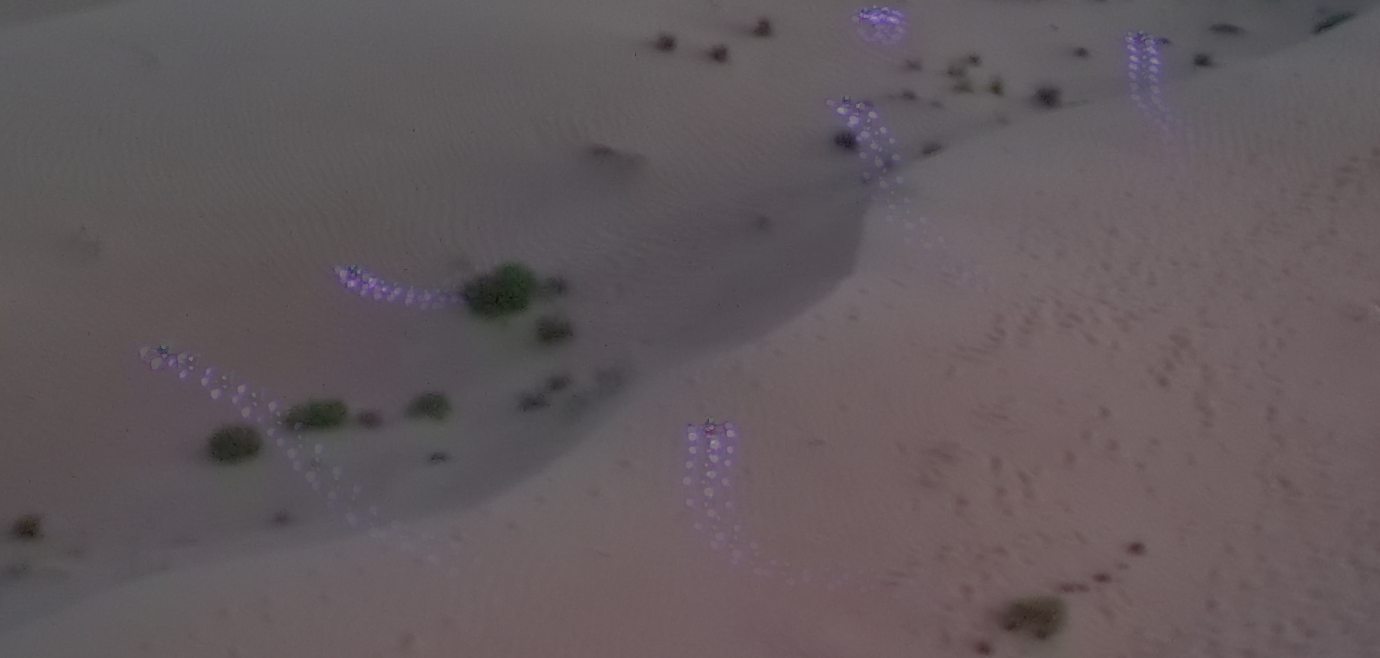}};

  \spy[width=1.45cm, height=2.5cm, magnification=3, 
    fill=none, 
    line width=1.pt,
    color=red,
    connect spies, 
     spy connection path={\draw[thick](tikzspyonnode) -- (tikzspyinnode);}]
    on (2.9, .6) in node[opacity=1, thick] at (7., 1.3);

\end{tikzpicture}

%% file: algorithm_overview.tex
\begin{tikzpicture}
    \tikzstyle{process} = [rectangle, rounded corners, draw=black, fill=orange!40, minimum width=4cm, minimum height=2cm, align=center]
    \tikzstyle{storage} = [rectangle, draw=teal, fill=teal!20, minimum width=4cm, minimum height=2cm, align=center]
    \tikzstyle{arrow} = [thick, ->, >=stealth]
    
    \draw[draw=teal, fill=teal!20, opacity=0.6] (-9,-1.25) rectangle ++(4,3);
    \draw[draw=teal, fill=teal!20, opacity=0.33] (-8.75,-1) rectangle ++(4,3);
    \node[draw=teal, fill=teal!20, text=black, minimum width=4cm, minimum height=3cm, align=center] (cameraFrame) at (-7,0) 
        {\Large Camera Frame};
    
    \draw[draw=gray!50, fill=gray!50, opacity=0.2] (-2.5,-4.5) rectangle ++(17.5,9);
    
    \node[process] (localSearch) at (0,0) {
    \Large{Local Search}\\[5pt]
    Check if $p_{t,k}$ is in \\ 
    \emph{Local Search} area \\
    of \emph{t-series} in \B{}. 
    };

    \node[process] (extendedSearch) at (6,0) {
    \Large {Extended Search}\\[5pt]
    Polynomial regression on \\ 
    \emph{t-series} with CI as search area.\\[5pt]
    Check if $p_{t,k}$ is in the CI of \emph{t-series}.}; 
    
    \node[storage] (processed) at (0,-3) {Add $p_{t,k}$ to end of $t_i$;\\ mark $t_i$ as processed};
    
    \node[storage] (processed_extended) at (6,-3) {Add $p_{t,k}$ to end of $t_i$;\\ mark $t_i$ as processed};
    
    \node[storage] (new_ts) at (6,3) {Insert $p_{t,k}$ to end of \B{} as new \ts{}};
    
    \node[process] (verify) at (12,0) {
        \Large {Verification}\\[5pt]
        Verify that only valid \emph{t-series} \\
        are in \B{}; For each unprocessed \\
        \emph{t-series}, insert ``off''-\ps{} \\
        (marker turned off).
    };
    
    \draw[arrow] (cameraFrame.east) -- node[midway, above]{\Large \shortstack{for each\\ $p_{t,k} \in P_t$}} (localSearch.west);
    
    \draw[arrow] (localSearch.east) -- node[midway, above]{\Large No} (extendedSearch.west);
    
    \draw[arrow] (extendedSearch.north) -- node[midway, right]{\Large No} (new_ts.south);
    
    \draw[arrow] (localSearch.south) -- node[midway, right]{\Large Yes} (processed.north);
    \draw[arrow] (extendedSearch.south) -- node[midway, right]{\Large Yes} (processed_extended.north);
    
    \draw[arrow] (extendedSearch.east) -- (verify.west);
    \draw[arrow] (verify.south) -- (12,-2.5);
    \node at (12,-3) {\shortstack{Extracted Blinking marker\\IDs with pixel locations}};

    \node at (-1.6,4) {\LARGE \textbf{\gls{AMT}}};

\end{tikzpicture}

%% file: buffer.tex
\begin{tikzpicture}[scale=1.2]
    \definecolor{gridblue}{RGB}{220,235,255}
    \definecolor{gridframe}{RGB}{70,130,180}
    \definecolor{filleddot}{RGB}{41,98,255}
    \definecolor{emptydot}{RGB}{255,255,255}
    \definecolor{timestamp}{RGB}{0,0,0}
    
    \fill[gridblue, rounded corners=3pt, opacity=0.5] (0,9) rectangle ++(14,5);
    \draw[gridframe, line width=1.5pt, rounded corners=3pt] (0,9) rectangle ++(14,5);
    
    \foreach \x in { 3.5, 5, 6.5, 8, 9.5, 11, 12.5}
        \draw[gridframe!60, fill=white, rounded corners=1pt] (\x,13) rectangle ++(1.5,1);

    \foreach \x in {2 , 3.5, 5, 6.5, 8, 9.5, 11, 12.5}
        \draw[gridframe!60, fill=white, rounded corners=1pt] (\x,11) rectangle ++(1.5,1);
    
    \foreach \x in {5, 6.5, 8, 9.5, 11, 12.5}
        \draw[gridframe!60, fill=white, rounded corners=1pt] (\x,9) rectangle ++(1.5,1);
    
    \foreach \y in {12.75, 12.5, 12.25} {
        \filldraw (13.5,\y) circle (0.8pt);
        \filldraw (-0.5,\y) circle (0.8pt);
    }
    
    \foreach \y in {10.75, 10.5, 10.25} {
        \filldraw (13.5,\y) circle (0.8pt);
        \filldraw (-0.5,\y) circle (0.8pt);
    }

    \foreach \x/\f in {4.25, 5.75/fill, 7.25/fill, 8.75/empty, 10.25/empty, 11.75/empty, 13.25/fill} {
        \draw[line width=1pt] (\x,13.5) circle (2pt);
    }
    \foreach \x/\f in {4.25/empty, 7.25, 10.25, 11.75/empty} {
        \filldraw[line width=1pt] (\x,13.5) circle (2pt);
    }

    \foreach \x/\f in {4.25/empty, 5.75/fill, 7.25/fill, 8.75/empty, 10.25/empty, 11.75/empty, 13.25/fill} {
        \draw[line width=1pt] (\x,11.5) circle (2pt);
    }
    \foreach \x/\f in {2.75, 4.25/empty, 8.75/empty, 11.75/empty} {
        \filldraw[line width=1pt] (\x,11.5) circle (2pt);
    }
    
    \foreach \x/\f in {5.75/fill, 8.75/empty, 11.75/empty} {
        \draw[line width=1pt] (\x,9.5) circle (2pt);
    }
    \foreach \x/\f in {7.25/fill,10.25/empty, 13.25/fill} {
        \filldraw[line width=1pt] (\x,9.5) circle (2pt);
    }
    
    \node[text=timestamp, font=\Large] at (0.2, 14.7) [align=left] {time};
    \node[text=timestamp, font=\Large] at (0.35, 14.3) [align=left] {stamp};
    
    \node[text=timestamp, font=\Large] at (-1.4,14.2) [align=left] {\ts{}};
    \draw[-, thick] (-.1,14.1) -- (-1.1,14.7);
    
    \node[font=\LARGE\bfseries, ] at (-0.5,13.5) {$S_{1}$};
    \node[font=\LARGE\bfseries, ] at (-0.5,11.5) {$S_{i}$};
    \node[font=\LARGE\bfseries, ] at (-0.5,9.5) {$S_{n}$};
    
    \node[font=\Large, text=timestamp] at (13.5,14.4) {$t-1$};
    \node[font=\Large, text=timestamp] at (8.75,14.4) {$t-4$};
    \node[font=\Large, text=timestamp] at (4.25,14.4) {$t-7$};
    \node[font=\Large, text=timestamp] at (14.5,14.4) {$t$};
    
    \foreach \x in {12.8, 12.6, 12.4} {
        \filldraw (\x,14.4) circle (0.8pt);
    }
    \foreach \x in {9.6, 9.8, 10} {
        \filldraw (\x,14.4) circle (0.8pt);
    }
    \foreach \x in {7.4, 7.6, 7.8} {
        \filldraw (\x,14.4) circle (0.8pt);
    }
    \foreach \x in {3.4, 3., 3.2} {
        \filldraw (\x,14.4) circle (0.8pt);
    }
    
    \draw[red, line width=2pt, rounded corners=1pt] (14,11) rectangle ++(1.5,1);
    \filldraw[red] (14.75,11.5) circle (2pt);
\end{tikzpicture}

%% file: buffer_detailed.tex
\begin{tikzpicture}[scale=0.95]
  \definecolor{signalColor}{RGB}{200, 200, 220}
  \definecolor{extendedColor}{RGB}{230, 245, 230}
  \definecolor{timelineColor}{RGB}{50, 50, 50}
  \definecolor{newPointColor}{RGB}{220, 60, 60}
  \definecolor{deletedColor}{RGB}{180, 180, 180}
  
  \node (ExtendedSearch) [rectangle, dashed, draw=green!70!black, line width=1pt,
  fill=extendedColor, rounded corners=3pt,
  opacity=0.7,
  minimum width=125.5mm, minimum height=2.55cm, anchor=south west] at (0.85,-0.8) {};
  
  \node (SignalMatcher) [rectangle, dashed, draw=blue!50!black, line width=1pt,
  fill=signalColor, rounded corners=3pt,
  opacity=0.7,
  minimum width=68mm, minimum height=3.1cm, anchor=south west] at (6.9,-1.5) {};
  
  \node[anchor=south, font=\large\sffamily] at (SignalMatcher.south) {used for Signal Matching with $D$}; 
  \node[font=\large\sffamily] at (3.5, -1.2) {used by \textit{Extended Search}};
  
  \draw[thick, ->, >=stealth] (0.85,0) -- (14.5,0) node[right, font=\large\sffamily]{time};
  
  \foreach \x/\y in {3.2/3.8, 6.2/6.8, 9.2/9.8} {
      \draw[thick, dotted] (\x,0) -- (\y,0);
      \draw[thick, dotted] (\x,1) -- (\y,1);
  }
  
  \foreach \x/\y in {0.85/3.2, 3.8/6.2, 6.8/9.2, 9.8/14.5} {
      \draw[thick] (\x,0) -- (\y,0);
  }
  
  \draw[thick, -] (0.9,-1.8) -- (14,-1.8);
  \draw (0.9,-1.9) -- (0.9,-1.7);    
  \draw (14,-1.9) -- (14,-1.7);
  \node[font=\large\sffamily] at (4.5, -2.2) {$L_S$};
  
  \draw[thick, -] (7,-2.2) -- (14,-2.2);
  \draw (7,-2.3) -- (7,-2.1);    
  \draw (14,-2.3) -- (14,-2.1);
  \node[font=\large\sffamily] at (10.5, -2.5) {$L_{\mathcal{D}}$};
  
  \foreach \x in {2, 5, 8, 11, 13} {
      \draw[thick] (\x,0.1) -- (\x,-0.1);
  }
  
  \foreach \x/\descr in {2/t-L_S, 5/t-j, 8/t-L_{\mathcal{D}}, 11/t-1, 13/t} {
      \node[font=\large\sffamily] at (\x,-0.4) {$\descr$};
  }
  
  \node[rectangle, draw=deletedColor, rounded corners=2pt, fill=white, 
  minimum width=2cm, minimum height=0.4, thick, text=deletedColor] (p0) at (-0.27,1) {$p(S_i)_{t-L_{S}-1}$};
  
  \node[rectangle, draw, rounded corners=2pt, fill=white, minimum width=2cm, minimum height=0.5, thick] (p1) at (1.85,1) {$p(S_i)_{t-L_{S}}$};
  
  \node[rectangle, draw, rounded corners=2pt, fill=white, minimum width=2cm, minimum height=0.2, thick] (p2) at (5,1) {$p(S_i)_{t-j}$};
  
  \node[rectangle, draw, rounded corners=2pt, fill=white, minimum width=2cm, minimum height=0.5, thick] (p3) at (8,1) {$p(S_i)_{t-L_{\mathcal{D}}}$};
  
  \node[rectangle, draw, rounded corners=2pt, fill=white, minimum width=2cm, minimum height=0.5, thick] (p4) at (11,1) {$p(S_i)_{t-1}$};
  
  \node[rectangle, draw=newPointColor, rounded corners=2pt, fill=white, minimum width=2cm, minimum height=0.5, thick, text=black] (p5) at (13,1) {$p(S_i)_t$};

  \draw[-{Triangle[width=10pt,length=5pt]}, line width=3.5pt, color=newPointColor](13, 2.2) -- (13, 1.4);
  \node[align=center, font=\sffamily] at (13,2.6) {
      \textbf{new inserted}\\
      \textbf{\ps{}}
  };
  
  \draw[-{Triangle[width=10pt,length=5pt]}, line width=3.5pt, color=deletedColor](0, 2.2) -- (0, 1.4);
  \node[align=center, font=\sffamily] at (0,2.6) {
      \textbf{deleted by}\\
      \textit{Verification}
  };
\end{tikzpicture}

%% file: extended_search.tex
\begin{tikzpicture}
    
    \definecolor{gridframe}{RGB}{70,130,180}
        
    \foreach \x in {2, 3.5, 5, 6.5, 8, 9.5}
        \draw[gridframe!60, fill=white, rounded corners=1pt] (\x,9) rectangle ++(1.5,1);
    \foreach \x in {2, 3.5, 5}
        \draw[gridframe!60, fill=white, rounded corners=1pt] (\x,7.5) rectangle ++(1.5,1);
        
    \foreach \x/\f in {5.75/empty, 7.25/fill, 10.25} {
        \draw[line width=1pt] (\x,9.5) circle (2pt);
    }
    \foreach \x/\f in {2.75/fill, 4.25/empty, 8.75/fill} {
        \filldraw[line width=1pt] (\x,9.5) circle (2pt);
    }
    
    \draw[stealth-, line width=0.4mm, blue!70!black, shorten >=2pt] (2.75,8.2) to[out=90, in=270] (2.75,9.3);
    \draw[stealth-, line width=0.4mm, blue!70!black, shorten >=2pt] (4.25,8.2) to[out=90, in=270] (4.25,9.3);
    \draw[stealth-, line width=0.4mm, blue!70!black, shorten >=2pt] (5.75,8.2) to[out=80, in=270] (8.75,9.3);
    
    \filldraw[black] (4.25,8) circle (2pt);
    \filldraw[black] (2.75,8) circle (2pt);
    \filldraw[black] (5.75,8) circle (2pt);
    \node [] at (5.75,7) {$w_{t-2}$};
    \node [] at (2.75,7) {$w_{t-6}$};
    \node [] at (4.25,7) {$w_{t-5}$};
    \node at (1.5,9.5) {$S_{i}$};
    \node [] at (1.5,8) {$S^{+}_{i}$};
    \node at (1.3,10.6) {time};
    \node at (1.4,10.2) {stamp};
    \node at (11.5,10.4) {{$t$}};
    \node at (10.25,10.4) {{$t-1$}};
    \node at (8.75,10.4) {{$t-2$}};
    \node at (7.25,10.4) {{$t-3$}};
    \node at (5.75,10.4) {{$t-4$}};
    \node at (4.25,10.4) {{$t-5$}};
    \node at (2.75,10.4) {{$t-6$}};
\end{tikzpicture}

%% file: extended_search_2.tex
\begin{tikzpicture}

    \draw [blue,ultra thick] (0,0) rectangle ++(8,5);

    \draw [black,] (-6,4) -- (1,-3);
    \draw [black,dashed] (-6,4) -- (-7,5);
    
    \draw [black,] (1,-3) -- (0.9,-2.6);
    \draw [black,] (1,-3) -- (0.4,-2.9);
    
    \draw [black,] (-6.2,4) -- (-5.8,4);
    \draw [black,] (-4.2,2) -- (-3.8,2);
    \draw [black,] (-2.2,0) -- (-1.8,0);
    \draw [black,] (-0.2,-2) -- (.2,-2);
    
    \node at (-.8,-2) 
      {\Huge $t$};
    \node at (-3.6,0) 
      {\Huge $t-2$};
    
    \node at (-5.6,2) 
      {\Huge $t-5$};
    \node at (-7.6,4) 
      {\Huge $t-6$};
    
    \draw [blue,opacity=0.5,very thin](-4,4) -- (-4,9);
    \draw [blue,opacity=0.5,very thin](4,9) -- (-4,9);
    \draw [blue,opacity=0.5,very thin,dashed](4,9) -- (4,4);
    \draw [blue,opacity=0.5,very thin,dashed](-4,4) -- (4,4);
    \draw [blue, opacity=0.75,thick](-2,2) -- (-2,7);
    \draw [blue, opacity=0.75,thick](6,7) -- (-2,7);
    \draw [blue, opacity=0.75,thick,dashed](6,7) -- (6,2);
    \draw [blue, opacity=0.75,thick,dashed](-2,2) -- (6,2);
    \draw [line width=0.1cm ,red, path fading=north, fading angle=0] plot [smooth, tension=0.5] coordinates { (1.15, 9.35 ) (1.15,4.35) (7.15,0.35)};
    \draw [blue, opacity=0.50, thick, fill] ( 0.6, 6.6 ) rectangle ++ (0.3,0.3);
    \draw [blue, opacity=0.75, thick, fill] ( 1, 4.2 ) rectangle ++ (0.3,0.3);
    \draw [blue, thick,fill] ( 4.4, 1.7 ) rectangle ++ (0.3,0.3);
    \draw [opacity=1,very thick, fill, red] ( 7, 0.2 ) rectangle ++ (0.3,0.3);
    \draw [opacity=1,very thick, red] ( 5.9, -0.5 ) rectangle ++ (2.4, 1.8);
    \draw [thick] (2,-2) rectangle ++(8,5);

\end{tikzpicture}

%% file: verification_buffer.tex
\begin{tikzpicture}[scale=1.2]
    \definecolor{gridblue}{RGB}{220,235,255}
    \definecolor{gridframe}{RGB}{70,130,180}
    \definecolor{filleddot}{RGB}{41,98,255}
    \definecolor{emptydot}{RGB}{255,255,255}
    \definecolor{timestamp}{RGB}{0,0,0}
    
    \fill[gridblue, rounded corners=3pt, opacity=0.5] (0,9) rectangle ++(14,5);
    \draw[gridframe, line width=1.5pt, rounded corners=3pt] (0,9) rectangle ++(14,5);
    
    \foreach \x in { 3.5, 5, 6.5, 8, 9.5, 11, 12.5, 14}
        \draw[gridframe!60, fill=white, rounded corners=1pt] (\x,13) rectangle ++(1.5,1);

    \foreach \x in {2 , 3.5, 5, 6.5, 8, 9.5, 11, 12.5}
        \draw[gridframe!60, fill=white, rounded corners=1pt] (\x,11) rectangle ++(1.5,1);
    
    \foreach \x in {5, 6.5, 8, 9.5, 11, 12.5, 14}
        \draw[gridframe!60, fill=white, rounded corners=1pt] (\x,9) rectangle ++(1.5,1);
    
    \foreach \y in {12.75, 12.5, 12.25} {
        \filldraw (13.5,\y) circle (0.8pt);
        \filldraw (-0.5,\y) circle (0.8pt);
    }
    
    \foreach \y in {10.75, 10.5, 10.25} {
        \filldraw (13.5,\y) circle (0.8pt);
        \filldraw (-0.5,\y) circle (0.8pt);
    }

    \draw[gridframe!60, fill=white, rounded corners=1.5pt] (14,11) rectangle ++(1.5,1);

    \foreach \x/\f in {4.25, 5.75/fill, 7.25/fill, 8.75/empty, 10.25/empty, 11.75/empty, 13.25/fill} {
        \draw[line width=1pt] (\x,13.5) circle (2pt);
    }
    \foreach \x/\f in {4.25/empty, 7.25, 10.25, 11.75/empty, 14.75} {
        \filldraw[line width=1pt] (\x,13.5) circle (2pt);
    }

    \foreach \x/\f in {4.25/empty, 5.75/fill, 7.25/fill, 8.75/empty, 10.25/empty, 11.75/empty, 13.25/fill, 14.75} {
        \draw[line width=1pt] (\x,11.5) circle (2pt);
    }
    \foreach \x/\f in {2.75, 4.25/empty, 8.75/empty, 11.75/empty, 13.25} {
        \filldraw[line width=1pt] (\x,11.5) circle (2pt);
    }
    
    \foreach \x/\f in {5.75/fill, 8.75/empty, 11.75/empty} {
        \draw[line width=1pt] (\x,9.5) circle (2pt);
    }
    \foreach \x/\f in {7.25/fill,10.25/empty, 13.25/fill, 14.75} {
        \filldraw[line width=1pt] (\x,9.5) circle (2pt);
    }
    
    \node[text=timestamp, font=\Large] at (0.2, 14.7) [align=left] {time};
    \node[text=timestamp, font=\Large] at (0.35, 14.3) [align=left] {stamp};
    
    \node[text=timestamp, font=\Large] at (-1.4,14.2) [align=left] {\ts{}};
    \draw[-, thick] (-.1,14.1) -- (-1.1,14.7);
    
    \node[font=\LARGE\bfseries, ] at (-0.5,13.5) {$S_{1}$};
    \node[font=\LARGE\bfseries, ] at (-0.5,11.5) {$S_{i}$};
    \node[font=\LARGE\bfseries, ] at (-0.5,9.5) {$S_{n}$};
    \node[font=\LARGE\bfseries, ] at (-0.5,8.5) {$S_{n+1}$};
    
    \node[font=\Large, text=timestamp] at (13.5,14.4) {$t-1$};
    \node[font=\Large, text=timestamp] at (8.75,14.4) {$t-4$};
    \node[font=\Large, text=timestamp] at (4.25,14.4) {$t-7$};
    \node[font=\Large, text=timestamp] at (14.5,14.4) {$t$};
    
    \foreach \x in {12.8, 12.6, 12.4} {
        \filldraw (\x,14.4) circle (0.8pt);
    }
    \foreach \x in {9.6, 9.8, 10} {
        \filldraw (\x,14.4) circle (0.8pt);
    }
    \foreach \x in {7.4, 7.6, 7.8} {
        \filldraw (\x,14.4) circle (0.8pt);
    }
    \foreach \x in {3.4, 3.6, 3.2} {
        \filldraw (\x,14.4) circle (0.8pt);
    }
    
    \fill[gray, , opacity=0.4] (3.49,13) rectangle ++(12,1);
    \fill[gray, , opacity=0.4] (5,9.) rectangle ++(10.5,1);
    \draw [ultra thick, red,-{Latex[length=2.8mm,width=3.5mm]} ](13.45,11.5) -- (14.6,11.5);
    \draw[green, fill=white, line width=2.5, rounded corners=1.5pt] (14,8) rectangle ++(1.5,1);
    \filldraw[line width=1pt] ( 14.75, 8.5) circle (2pt);
\end{tikzpicture}

%% file: outdoor_2.tex
\begin{tikzpicture}
    \node[anchor=south west,inner sep=0] (a) at (0,0) { 
    \includegraphics[width=0.45\textwidth]{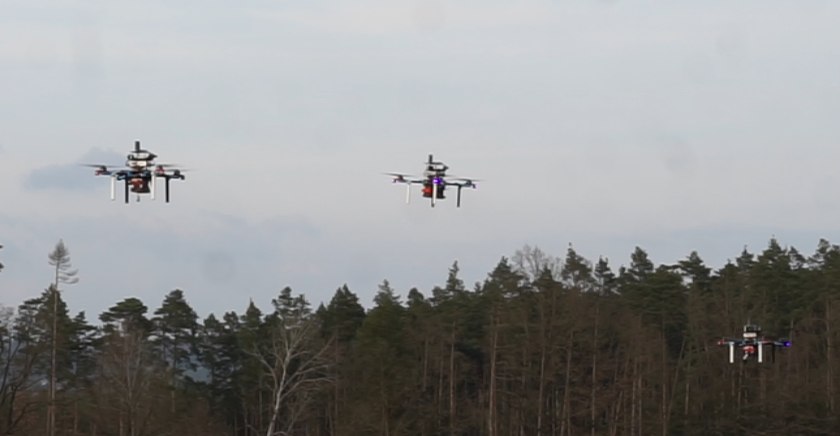}};
    
    \begin{scope}[x={(a.south east)},y={(a.north west)}]

      \node [black] at(0.1, 0.5) {\glsentryshort{RX}};
      \node [black] at(0.5, 0.48) {\glsentryshort{TX}1};
      \node [white] at(0.82, 0.12) {\glsentryshort{TX}2};

    \end{scope}
  \end{tikzpicture}

%% file: experiment_setup.tex
\begin{tikzpicture}[decoration = {markings,
  mark = at position 0.0 with {\arrow[>=stealth]{<}}}]
  \draw[step=1cm,lightgray,very thin] (-4.2,-4.2) grid (9.2,4.2);

  \newdimen\R     
  \R=2cm
  \draw[-stealth', blue, dashed] (-150:{0.4*\R}) arc (-150:150:{0.4*\R});

  \draw [line width=0.3mm, -] (-3,-4) -- (-4,-4.) ;
  \draw [line width=0.3mm, -] (-3,-3.85) -- (-3,-4.15) ;
  \draw [line width=0.3mm, -] (-4,-3.85) -- (-4,-4.15) ;
  \node at (-3.5,-3.7) {\large 1m};

  \draw [dashed, ultra thick,<->, orange] (0,0) -- (-4,0);

  \draw [dashed, <->, ultra thick,teal] (4,4) -- (4,-4);
  \draw [dashed, <->, ultra thick,teal] (8,4) -- (8,-4);
  \draw [dashed, <->, ultra thick,magenta] (0,4) -- (0,-4);

  \node [orange] at (-2,0.2) {\large Exp. 3};
  \node [blue] at (1,1.1) {\large Exp. 1};
  \node [magenta] at (-0.8,-2.7) {\large Exp. 2};
  \node [magenta] at (3.2,-2.7) {\large Exp. 2};
  \node [magenta] at (7.2,-2.7) {\large Exp. 2};
  \node [olive] at (3.2,-3.7) {\large Exp. 7};
  \node [teal] at (3.2,-3.2) {\large Exp. 4};
  \node [teal] at (7.2,-3.2) {\large Exp. 4};
   
  \pic [rotate=0, transform shape, scale=0.4] at (4.0,0.0) {tx};
  \pic [rotate=0, transform shape, scale=0.4] at (8.0,1.0) {tx};
  \pic [rotate=-15, transform shape, scale=0.4] at (0.0,0.0) {observer};

  \node []at (4.6, 0.6) {\large 0};
  \node []at (4.6, -0.6) {\large 3};
  \node[] at (3.4, -0.6) {\large 2};
  \node []at (3.4, 0.6) {\large 1};
  
  \node []at (8.6, 1.6) {\large 4};
  \node []at (8.6, 0.4) {\large 7};
  \node[] at (7.4, 0.4) {\large 6};
  \node []at (7.4, 1.6) {\large 5};
  
  \node at (4.8, 0.0) {\Large\glsentryshort{TX}1};
  \node at (8.8, 1.) {\Large\glsentryshort{TX}2};
  \node at (-1.2, -0.8) {\Large\glsentryshort{RX}};

  \draw [line width=0.3mm, -stealth] (0,0) -- (1.5,0);
  \draw [line width=0.3mm, -stealth] (0,0) -- (0,1.5) ;
  \node at (-0.2,1.3) {\large y};
  \node at (1.3,-0.2) {\large x};
\end{tikzpicture}

%% file: star_setup.tex
\begin{tikzpicture}[decoration = {markings,
  mark = at position 0.0 with {\arrow[>=stealth]{<}}
}
]
\draw[step=1cm,lightgray,very thin] (-2.75,-0.2) grid (2.75,6.2);

  \draw [line width=0.3mm, -] (1,0) -- (2.,0) ;
  \draw [line width=0.3mm, -] (1,-.15) -- (1,0.15) ;
  \draw [line width=0.3mm, -] (2,-0.15) -- (2,0.15) ;
  \node at (1.5,0.2) {\large 1m};

  \filldraw [] (0, 6) circle (1pt);
  \filldraw [] (2, 3) circle (1pt);
  \filldraw [] (-2, 3) circle (1pt);
  \filldraw [] (-2.5, 5.5) circle (1pt);
  \filldraw [] (2.5, 5.5) circle (1pt);

  \draw [line width=0.3mm, -{Latex[length=2mm,width=3mm]}] (0,6) -- (-2,3.);
  \draw [line width=0.3mm, {Latex[length=2mm,width=3mm]}-] (2.5,5.5) -- (-2,3.) ;
  \draw [line width=0.3mm, -{Latex[length=2mm,width=3mm]}] (2.5,5.5) -- (-2.5,5.5) ;
  \draw [line width=0.3mm, {Latex[length=2mm,width=3mm]}-] (2,3) -- (-2.5,5.5) ;
  \draw [line width=0.3mm, -{Latex[length=2mm,width=3mm]}] (2,3) -- (0,6) ;

  \draw [line width=0.3mm, -stealth] (0,0) -- (-1.5,0);
  \draw [line width=0.3mm, -stealth] (0,0) -- (0,1.5) ;
  \node at (-0.2,1.3) {\large z};
  \node at (-1.3,-0.3) {\large y};
\end{tikzpicture}